\documentclass{article}

\usepackage{arxiv}

\usepackage[utf8]{inputenc} 
\usepackage[T1]{fontenc}    
\usepackage{hyperref}       
\usepackage{url}            
\usepackage{amsfonts}       
\usepackage{nicefrac}       
\usepackage{microtype}      
\usepackage{graphicx}
\usepackage{doi}
\usepackage{caption}

\usepackage{booktabs}  
\usepackage{chngpage}
\usepackage{amsmath}
\usepackage{threeparttable}
\usepackage{siunitx}
\graphicspath{ {./figures/} }
\usepackage{cleveref}

\usepackage[acronym]{glossaries}
\newacronym{cp}{CP}{critical power}
\newacronym{w'}{\ensuremath{W^\prime}}{finite energy reserve for work above critical power}
\newacronym{w'bal}{\ensuremath{W^\prime_{bal}}}{\ensuremath{W^\prime} balance}
\newacronym{tte}{TTE}{time to exhaustion}

\title{A New Pathway to Approximate Energy Expenditure and Recovery of an Athlete}

\date{Sept 13, 2021}

\author{
  \href{https://orcid.org/0000-0001-8868-9735}{\includegraphics[scale=0.06]{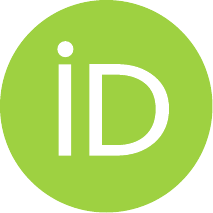}\hspace{1mm}}Fabian Clemens Weigend\\
  School of Computer, Data and Mathematical Sciences and
  School of Health Sciences\\
  Western Sydney University\\
  Australia\\
  \texttt{Fabian.Weigend@westernsydney.edu.au} \\
   \And
  \href{https://orcid.org/0000-0003-1346-4982}{\includegraphics[scale=0.06]{orcid.pdf}}\hspace{1mm}Jason Siegler\\
  College of Health Solutions\\
  Arizona State University\\
  USA\\
  \texttt{Jason.Siegler@asu.edu} \\
  \And
  \href{https://orcid.org/0000-0002-8284-2062}{\includegraphics[scale=0.06]{orcid.pdf}}\hspace{1mm}Oliver Obst \\
  School of Computer, Data and Mathematical Sciences\\
  Western Sydney University\\
  Australia \\
  \texttt{O.Obst@westernsydney.edu.au} \\
}

\begin{document}
\maketitle

\begin{abstract}
This work proposes to use evolutionary computation as a pathway to allow a new perspective on the modeling of energy expenditure and recovery of an individual athlete during exercise. 

We revisit a theoretical concept called the ``three component hydraulic model'' which is designed to simulate metabolic systems during exercise and which is able to address recently highlighted shortcomings of currently applied performance models. This hydraulic model has not been entirely validated on individual athletes because it depends on physiological measures that cannot be acquired in the required precision or quantity.

This paper introduces a generalized interpretation and formalization of the three component hydraulic model that removes its ties to concrete metabolic measures and allows to use evolutionary computation to fit its parameters to an athlete.
\end{abstract}

\keywords{Performance Modeling \and Metabolic Response Modeling \and Optimization}

\section{Introduction}
\label{sec:introduction}

The research area of performance modeling can be considered as the generalization of physiological processes into  mathematical  models  with  the  purpose  of  approximating  a  body’s  response to exercise. Created models represent an objective understanding of body responses and provide opportunities to serve as reasoning tools to be applied in performance prediction, training simulation or exercise prescription~\cite{clarke_rationale_2013}. Jones and Vanhatalo~\cite{jones_critical_2017} as well as Sreedhara et al.~\cite{sreedhara_survey_2019} agree, that current approaches require refinement but highlight their potential for future research opportunities.  

With their in 2019 published findings, Caen et al.~\cite{caen_reconstitution_2019} propose that recovery kinetics depend on previous work rate in addition to time and exercise intensity during recovery. Established performance modeling approaches separate energy expenditure and recovery into two separate models \cite{clarke_rationale_2013, jones_critical_2017, sreedhara_survey_2019} and are too restricted to address these findings. 

This work proposes that evolutionary computation allows to revisit an alternative concept: The so-called ``three component hydraulic model'' is investigated by Morton~\cite{morton_three_1986, morton_critical_2006} and represents human energy systems as three interconnected tanks. Energy is represented by liquids, the flow out of a tap corresponds to energy expenditure, and the refilling of tanks can be understood as recovery. Its conceptualization allows to combine expenditure and recovery in one model and recent findings by Caen et al.~\cite{caen_reconstitution_2019} can be addressed. However, Morton still highlights that the model is not absolute realistic in its assumptions and that more---and more precise---physiological measurements are needed to see how many predictions of this model conform to reality~\cite{morton_critical_2006}. 

We introduce a generalized form of the three component hydraulic model that removes ties to such concrete metabolic measures and opens up the opportunity to use evolutionary computation as a new pathway to apply it to individual athletes.

\section{A Pathway to Apply The Hydraulic Performance Model}
\label{sec:pathway_to_apply_model}

To enable the application of evolutionary computation, the three component hydraulic system is further generalized. In short, concrete relations to lactate, carbohydrate or phosphocreatine are removed. The three interacting components of the model are instead understood as more abstract entities and referred to as the anaerobic fast component ($AnF$), the anaerobic slow component ($AnS$), and the aerobic contribution ($Ae$). Model equations are developed in accordance to approaches by~\cite{morton_three_1986}. Detailed equations and more information about the generalization are provided in Appendix A.

A configuration of our generalized three component hydraulic model consists of eight parameters, which affect how the model's three components ($AnF$, $AnS$, $Ae$) interact. To fit such a configuration of eight parameters to an athlete, two objectives are defined: One for energy expenditure and one for energy recovery.

For the energy expenditure objective, the so-called ``critical power concept'' is employed as the ground truth. It is the established model to estimate times to exhaustion and used in most of the currently applied performance models~\cite{clarke_rationale_2013, jones_critical_2017, sreedhara_survey_2019}. A total of 12 performance tests at a constant exercise intensity until exhaustion are simulated by the generalized three component hydraulic model. The normalized root mean squared error of all differences between simulated and expected times to exhaustion is to be minimized as the expenditure objective.

The energy recovery objective uses the exercise protocol that Caen et al.~\cite{caen_reconstitution_2019} employed to obtain their published recovery ratios. The generalized hydraulic model simulates all trials that were conducted by Caen et al., which results in 12 differences between simulated and expected recovery ratios. Also here the normalized root mean squared error of all differences is used to determine the energy recovery fitness measure to be minimized. More information on the exercise protocols and corresponding simulations is provided in Appendix B.

To find a configuration for the hydraulic model that optimizes both objectives, the established Multi-Objective Evolutionary Algorithm with Decomposition (MOEA/D) approach coupled with the asynchronous islands functionality of Pygmo~\cite{biscani_parallel_2020} is used. Since both objective functions result in a normalized root mean squared error of 12 measurements, they are directly comparable. Knowledge of the evolved Pareto front allows to derive the best trade-off between both dynamics as the configuration that has the smallest Euclidean distance to the minimal error, i.e., point $(0,0)$. If more details on this rationale are needed, please see Appendix C.

We want to emphasize that the introduced approach serves as a proof of concept and much more room for parameter optimization and exploration of problem-specific algorithms is left for future work. Nearly all parameters are at the default that Pygmo provides and only four parameters were investigated by grid search. More details are given in Appendix C. 

\section{Results And Discussion}
\label{sec:results}

Ten independently estimated evolutionary fittings with the described hydraulic model and algorithm parameters are investigated and compared regarding consistency and quality.

\begin{figure}
    \centering
    \captionsetup{width=0.65\linewidth}
    \includegraphics[width=0.65\linewidth]{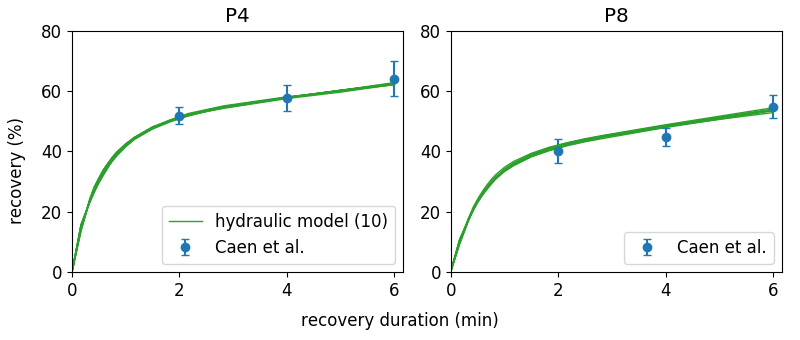}
    \caption{Recovery dynamics of fitted hydraulic models in comparison to published observations by Caen et al.~\cite{caen_reconstitution_2019}.
    }
    \label{fig:hyd_rec_compare}
\end{figure}

\begin{figure}
    \centering
    \captionsetup{width=0.65\linewidth}
    \includegraphics[width=0.65\linewidth]{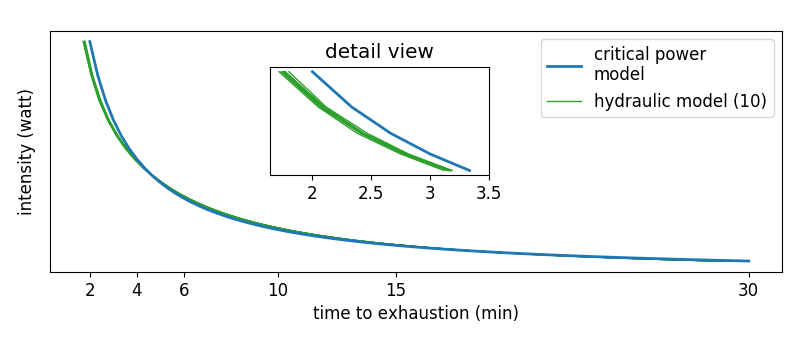}
    \caption{Exercise to time to exhaustion relationship of fitted hydraulic models in comparison to the critical power model.}
    \label{fig:hyd_exp_compare}
\end{figure}

All ten results behave similar to each other. As observable in \Cref{fig:hyd_exp_compare}, the---by the critical power concept suggested---hyperbolic relationship of exhaustive exercise intensity and time to exhaustion is closely recreated with only slight deviations in the high intensities. Also simulated recovery ratios after various conditions are similar to the ones observed by Caen et al.~\cite{caen_reconstitution_2019}. The values P4 and P8 represent preceding exhaustive work bout intensities, i.e., P4 is the intensity that is predicted to lead to exhaustion after 4 min. Published means and standard deviations for P4 (2 min : $51.8\% \pm 2.8\%$, 4 min: $57.7\% \pm 4.3\%$, 6 min: $64\% \pm 5.8\%$) and P8 ($40.1\% \pm 3.9\%$, $ 44.8\% \pm 3\%$, $54.8\% \pm 3.8\%$) are denoted in \Cref{fig:hyd_rec_compare}. 

Overall, evolved configurations make the hydraulic model successfully resemble the hyperbolic intensity to time to exhaustion relationship (\Cref{fig:hyd_exp_compare}) as well as recovery ratios that are affected by previous energy expenditure characteristics (\Cref{fig:hyd_rec_compare}). The outlined evolutionary computation approach successfully allows to apply three component hydraulic model as a performance model that addresses findings by Caen et al.~\cite{caen_reconstitution_2019} and combines expenditure and recovery in one concept. 

\section{Conclusion}
\label{sec:conclusion}

The proposed evolutionary computation pathway and generalized understanding of the three component hydraulic model add a beneficial new perspective to research in performance modeling. Results clearly motivate further investigations as a validation strategy for the hydraulic concept and to bring the model closer to the application on individual athletes. 

The implemented hydraulic model and evolutionary approach are available at \url{https://github.com/faweigend/three_comp_hyd}.

\section*{ACKNOWLEDGMENTS}
We thank Markus Wagner (University of Adelaide) for his support and feedback.

\appendix

\section{Model Generalization and Formalization}

Our generalized interpretation of the three component hydraulic model removes its ties to concrete physiological measures. This section defines this more general view with a robust formalization of its dynamics. These steps---generalization and formalization---ultimately allow to see the fitting of the three component hydraulic model as a two-objective optimization problem with eight parameters that can be approached with evolutionary computation.

\subsection{Model Generalization}

\begin{figure}
    \centering
    \captionsetup{width=0.6\linewidth}
    \includegraphics[width=0.6\linewidth]{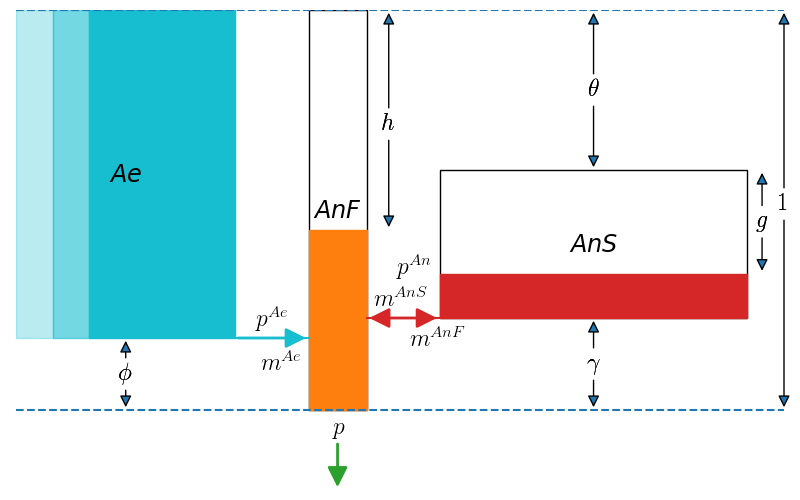}
    \caption{A generalized schematic of the three component hydraulic model. Tanks are renamed as aerobic component $Ae$, anaerobic fast component $AnF$ and anaerobic slow component $AnS$. $p^{Ae}$ and $p^{An}$ represent flows and $- m^{AnF}$, $m^{AnS}$, and $m^{Ae}$ maximal flow capacities.}
    \label{fig:hyd_three_comp_update}
\end{figure}

A schematic of the generalized form of the three component hydraulic model is depicted in~\Cref{fig:hyd_three_comp_update}. Ignoring relations to lactate, carbohydrate or phosphocreatine, this work refers to the middle tank ($A_nA$ in Morton's schematic in Figure 5 of his review from 2006~\cite{morton_critical_2006}) as the anaerobic fast component $AnF$ and the tank on the right ($A_nL$ in Figure 5 of Morton's review \cite{morton_critical_2006}) as the anaerobic slow component $AnS$. The left tank originally labeled with $O$ for oxygen is renamed into the aerobic contribution $Ae$. This more general interpretation also allows to fully remove tube $B$, which was included in Morton's work to account for early lactate levels in blood~\cite{morton_three_1986}.

These are just slight adjustments but they represent a new perspective on the dynamics of the model. Rather than the originally intended concrete metabolic energy storages that have to be set up according to empirical measures, tanks now represent more abstract entities that allow to be interpreted as a combination of sources which can be fitted by optimizing an objective function.

\subsection{Configurations}

A configuration of our proposed adjusted three component hydraulic model entails component positions, sizes and capacities. In alignment to \Cref{fig:hyd_three_comp_update}, a configuration $c$ for the model is defined as a collection of the following values
\begin{equation} \label{eq:configuration}
    c = \langle AnF, AnS, m^O, m^{AnS}, m^{AnF}, \phi, \theta, \gamma \rangle,
\end{equation}
where $\lbrace AnF,AnS \rbrace$ are tank capacities, $\lbrace  m^O, m^{AnS}, m^{AnF} \rbrace$ are maximal flow capacities and $\lbrace \phi, \theta, \gamma \rbrace$ are distances to define tank and pipe positions.

\subsection{Model Formalization}
\label{app:extended_formalization}

For simulations that allow to find an optimal configuration $c$ via an evolutionary computation approach, the model needs to be formalized in a robust manner. All equations are detailed in correspondence to the notation depicted in \Cref{fig:hyd_three_comp_update} and in accordance to approaches by Morton~\cite{morton_modelling_1990, morton_model_1985, morton_model_1986} and Sundstr\"om~\cite{sundstrom_comparing_2014} who build upon Morton's work. Our simulations do not include efforts where the athlete has to work at the maximal intensity they can possibly sustain and therefore we do not cover Morton's and Sundstr\"om's limitations on maximal power output (limitations for a maximal $p$).

A simulation starts with the drainage of liquid to match power demands $p$. The simulation uses discrete time steps and the time difference between two time steps $t$ and $t + 1$ is denoted as $\Delta t$. For the estimations of fill levels and flows for time step $t$, first the previous $h_{t-1}$ is adapted according to the power demand $p_t$. This results in the intermediate level
\begin{equation}\label{eq:h_p_t}
    h^p_t = h_{t-1} + \frac{p_t}{AnF} \cdot \Delta t.
\end{equation}
Now the liquids in tanks $Ae$ and $AnS$ react to the new fill level of $AnF$ and flows are estimated. The contribution $p^{Ae}$ from the $Ae$ tank is estimated as follows
\begin{equation}\label{eq:p_ae}
  p^{Ae}_t = 
  \begin{cases}
    m^{Ae} \cdot \frac{h^p_t}{1-\phi}, & \text{if $0 \leq h^p_t \leq (1 - \phi$)}.\\
    m^{Ae}, & \text{otherwise}.
  \end{cases}
\end{equation}
The maximal possible contribution $m^{Ae}$ is scaled with the ratio of the fill level of $AnF$ to $(1-\phi)$, which means the maximal flow is reached as soon as $h^p_t \geq (1-\phi)$. Because the size of $Ae$ is infinite, liquid will never flow back into $Ae$ and thus the interval of $p^{Ae}$ is $[0, m^{Ae}]$.

Estimations of the flow from $AnS$ to $AnF$ or backwards from $AnF$ to $AnS$ are more sophisticated. The flow through this pipe is defined as $p^{An}$ and, because liquid can refill $AnS$ or flow out of it, the interval is $[-m^{AnF}, m^{AnS}]$. Let $g^{max}$ be defined as the total height of $AnS$: 
\begin{equation}\label{eq:g_max}
    g^{max} = 1 - \theta - \gamma.
\end{equation}
To introduce possible flows more clearly, calculations are introduced in categories. The full \Cref{eq:p_an_total} for $p^{An}_t$, is the combination of \Cref{eq:p_an=0}, \Cref{eq:p_an=+}, and \Cref{eq:p_an=-} of this Appendix. \Cref{eq:p_an=0} describes cases in which no flow between $AnS$ and $AnF$ happens and thus $p^{An}_t$ equals 0:
\begin{equation}\label{eq:p_an=0}
  p^{An}_t = 
  \begin{cases}
    0, & \text{if $h^p_t \leq \theta$}\\
    & \text{and $g_{t-1} = 0$}.\\
    0, & \text{if $h^p_t \geq (1-\gamma)$}\\
    & \text{and $g_{t-1} = g^{max}$}.\\
    0, & \text{if $h^p_t = (g_{t-1} + \theta)$}.
  \end{cases}
\end{equation}
In the first case, the tank $AnS$ is full and the fill level of $AnF$ is above the top of tank $AnS$. In the second case, the fill level of $AnF$ is below the bottom end of $AnS$ and $AnS$ is empty. Finally, in the third one, the fill level of $AnF$ is exactly at par with the fill level of $AnS$ causing an equilibrium between both.

In \Cref{eq:p_an=+} cases in which liquid flows out of $AnS$ into $AnF$ are covered:
\begin{equation}\label{eq:p_an=+}
  p^{An}_t = 
  \begin{cases}
    m^{AnS} \cdot \frac{h^p_t-(g_{t-1} + \theta)}{g^{max}}, & \text{if $h^p_t > (g_{t-1} + \theta)$} \\ 
    & \text{and $h^p_t<(1-\gamma)$}.\\
    m^{AnS} \cdot \frac{g^{max} - g_{t-1}}{g^{max}}, & \text{if $h^p_t \geq (1 - \gamma) $} \\
    & \text{and $g_{t-1} < g^{max}$}.
  \end{cases}
\end{equation}
If the fill level of $AnF$ is below the fill level of $AnS$ and above the bottom end of $AnS$, the maximal possible flow is scaled according to the ratio of the difference between fill levels and the total height of $AnS$. Or, if the fill level of $AnF$ is below the bottom end of $AnS$ and $AnS$ is not empty, the maximal flow is scaled according to the amount of remaining liquid to consider the pressure of remaining liquid in the tank. 

\Cref{eq:p_an=-} describes the refilling flow---the flow back from $AnF$ into $AnS$:
\begin{equation}\label{eq:p_an=-}
  p^{An}_t = 
  \begin{cases}
    m^{AnF} \cdot \frac{h^p_t-( g_{t-1} + \theta)}{1-\gamma}, & \text{if $h^p_t < (g_{t-1} + \theta)$} \\
    & \text{and $g_{t-1} > 0$}.\\
  \end{cases}
\end{equation}
Here the fill level of $AnF$ is above the fill level of $AnS$ and $AnS$ is not full, which causes liquid to flow back into $AnS$. The maximal flow $m^{AnF}$ from $AnF$ into $AnS$ is scaled according to the ratio between the difference of fill levels and the height of $AnS$. Since $h^p_t$ is smaller than $g_{t-1} + \theta$, the result will be negative, indicating that a re-flow into $AnS$ happens. 

As the result, the full equation for $p^{An}_t$ is the combination of \Cref{eq:p_an=0}, \Cref{eq:p_an=+} and \Cref{eq:p_an=-}:
\begin{equation}\label{eq:p_an_total}
  p^{An}_t = 
  \begin{cases}
    0, & \text{if $h^p_t \leq \theta$}\\
    & \text{and $g_{t-1} = 0$}.\\
    0, & \text{if $h^p_t \geq (1-\gamma)$}\\
    & \text{and $g_{t-1} = g^{max}$}.\\
    0, & \text{if $h^p_t = (g_{t-1} + \theta)$}.\\
    m^{AnS} \cdot \frac{h^p_t-(g_{t-1} + \theta)}{g^{max}}, & \text{if $h^p_t > (g_{t-1} + \theta)$} \\ 
    & \text{and $h^p_t<(1-\gamma)$}.\\
    m^{AnS} \cdot \frac{g^{max} - g_{t-1}}{g^{max}}, & \text{if $h^p_t \geq (1 - \gamma) $} \\
    & \text{and $g_{t-1} < g^{max}$}.\\
    m^{AnF} \cdot \frac{h^p_t-( g_{t-1} + \theta)}{1-\gamma}, & \text{if $h^p_t < (g_{t-1} + \theta)$} \\
    & \text{and $g_{t-1} > 0$}.\\
  \end{cases}
\end{equation}
Having both flow values $p^{Ae}_t$ and $p^{An}_t$ of the current time step $t$, the tank fill levels of this time step are derived as:
\begin{align}
\begin{split}\label{eq:h_t}
    h_{t} = h^p_t - \frac{p^{An}_t + p^{Ae}_t}{AnF} \cdot \Delta t,
\end{split}\\
\begin{split}\label{eq:g_t}
    g_{t} = g_{t-1} + \frac{p^{An}_t}{AnS} \cdot \Delta t.
\end{split}
\end{align}
These equations allow to estimate tank fill levels for each time step $t$ throughout a simulation with possibly varying power demands $p_t$. 

\subsection{Handling Extreme Cases}

Large values for $\Delta t$, or combinations of small parameter values for tanks sizes with large values for maximal flow rates can cause faulty estimations for $p^{An}_t$.
These scenarios are not considered by Sundstr\"om~\cite{sundstrom_optimization_2013, sundstrom_comparing_2014} or Morton~\cite{morton_model_1985, morton_model_1986, morton_three_1986, morton_modelling_1990} because Morton used differential equations on isolated test scenarios and Sundstr\"om's handcrafted simulation conditions seem to not cause situations in which such extreme values come into effect.

In order to make simulations robust in such cases, three limitations to the flow $p^{An}$ are applied: In case $\Delta t$, $m^{AnF}$, or $m^{AnS}$ are large, it can occur that $p^{An}$ becomes larger than the remaining capacity of $AnS$ or the negative $p^{An}$ refills more liquid than $AnS$ can store. For the case that not enough is remaining in $AnS$, $p^{An}$ is capped to the remaining amount:
\begin{equation}\label{eq:cap_bottom_p_an}
    p^{An}_t \cdot \Delta t = \text{min}(p^{An}_t \cdot \Delta t, (g^{max}-g_{t-1}) \cdot AnS).
\end{equation}
Similarly, if $p^{An}$ amounts to more re-flow into $AnS$ than the available capacity, it is set to just fill $AnS$ to the top:
\begin{equation}\label{eq:cap_top_p_an}
    p^{An}_t \cdot \Delta t = \text{max}(p^{An}_t \cdot \Delta t, -g_{t-1} \cdot AnS).
\end{equation}
Further, large $\Delta t$, $m^{AnF}$ or $m^{AnS}$ as well as $AnF$ or $AnS$ capacities can cause $p^{An}$ to force a flow that overshoots the targeted equilibrium between both tank fill levels. Thus, the maximal flow $m^{An}_t$ between both tanks is defined and limits $p^{An}$:
\begin{equation}\label{eq:p_an_max}
m^{An}_t = \frac{h^p_t - (g_{t-1} + \theta)}{\frac{1}{AnS} + \frac{1}{AnF}}.
\end{equation}
Since both $p^{An}$ and $m^{An}_t$ may be negative or positive, the limitation applies in the following manner:
\begin{equation}\label{eq:p_an_max_app}
  p^{An}_t \cdot \Delta t = 
  \begin{cases}
    \text{max}(m^{An}_t,p^{An}_t \cdot \Delta t), & \text{if $p^{An}_t < 0$.}\\
    \text{min}(m^{An}_t,p^{An}_t \cdot \Delta t), & \text{if $p^{An}_t > 0$.}\\
  \end{cases}
\end{equation}
Applying these limits, model simulations stay robust even with extreme values and the model is ready to be fitted by an optimizer.

\section{Objective Functions}
\label{app:ground_truth_obj_function}

Two objectives capture how well a configuration $c$ for the hydraulic model makes it recreate measured performances of a tested athlete. In this context the term fitness then describes how well the model reproduces the expected responses. Two fitness measures are to be optimized: energy expenditure and recovery. The ground truth performance measures are an athlete's \gls*{w'} and \gls*{cp} as well as the group averaged recovery ratios that Caen et al.~\cite{caen_reconstitution_2019} observed on their participants. 

\subsection{Energy Expenditure}

For the energy expenditure objective function, time to exhaustion estimations of the critical power concept \cite{hill_critical_1993} are compared to time to exhaustion estimations of the hydraulic model. Using the critical power concept, exercise intensities can be derived that lead to exhaustion after a given amount of seconds. A total of 12 intensities are estimated for the energy expenditure fitness. These intensities are the ones that are estimated to lead to exhaustion after 120, 130, 140, 150, 170, 190, 210, 250, 310, 400, 600, 1200 seconds. The three component hydraulic model with configuration $c$ simulates constant exercise at these intensities. As soon as liquid flow out of the hydraulic model's tap $p$ cannot sustain the demand anymore, exhaustion is reached and the total time to exhaustion is compared to the expected one. From these trials, a total of 12 differences between the expected amount of seconds until exhaustion and the simulated amount of seconds until exhaustion are derived. The normalized root mean squared difference of these is the error measurement to be minimized for the expenditure objective.

\subsection{Energy Recovery}

For an estimation of energy recovery capabilities of a hydraulic model, the recovery ratios summarized in~\Cref{tab:recovery_ratio} are used. The three component hydraulic model with configuration $c$ simulates the same exercise protocol that Caen et al.~\cite{caen_reconstitution_2019} conducted to obtain their measurements.

Using the critical power concept \cite{hill_critical_1993}, work rates that lead to theoretical exhaustion after 4 min (P4) and 8 min (P8) are derived. Using these intensities Caen et al. obtained the in \Cref{tab:recovery_ratio} summarized recovery rates with a test setup that will be referred to as a work bout 1 (WB1) $\rightarrow$ recovery bout (RB) $\rightarrow$ work bout 2 (WB2) structure. It is conducted as follows: They let an athlete perform the first work bout (WB1) at a constant exercise intensity (in this case either P4 or P8) until the athlete cannot maintain this intensity anymore. This then assumes that \gls*{w'} is depleted. Afterwards immediately the recovery bout (RB) is started in which they switch to a much lower recovery intensity (in this case either 33\% of \gls*{cp} (CP33) or 66\% of \gls*{cp} (CP66)). This recovery phase prolongs for 2, 4 or 6 minutes and is followed by the second work bout (WB2) at the same intensity level as WB1 was conducted at. This second work bout also is stopped when the athlete cannot maintain the intensity anymore and the duration, i.e., \gls*{tte}, are recorded for both work bouts. Because of the very limited recovery bout duration in-between both work bouts, the \gls*{tte} of WB2 is bound to be shorter than the one of WB1, and the difference between both is considered to be the amount of \gls*{w'bal} that was reconstituted during the RB. 

As an example using \Cref{tab:recovery_ratio}, the first observation of the P4-CP33 line at 2 min represents a WB1 $\rightarrow$ RB $\rightarrow$ WB2 trial at the intensities P4 $\rightarrow$ CP33 $\rightarrow$ P4, where the recovery bout prolonged for 2 min. The duration of WB2 was 55\% of the duration of WB1 and thus it is inferred that the athlete could retain 55\% of their energy.

The three component hydraulic model with configuration $c$ conducts the same protocol and exhaustion is the point where liquid flow out of tap $p$ cannot meet the demand anymore. The normalized root mean squared error of these resulting 12 differences between observed recovery ratios and simulated recovery ratios is used as the error to be minimized for the recovery objective.

\begin{table}
\centering
\caption{Recovery ratios derived from~\cite{caen_reconstitution_2019}}
\label{tab:recovery_ratio}
\begin{tabular}{ l|c|c|c }
 & 2 min & 4 min & 6 min \\
\hline
    P4 - CP33 & 55 \% & 61 \% & 70.5 \% \\
    P4 - CP66 & 49 \% & 55 \% & 58 \% \\
    P8 - CP33 & 42 \% & 52 \% & 59.5 \% \\
    P8 - CP66 & 38 \% & 37.5 \% & 50 \% \\
\end{tabular}
\begin{tablenotes}
  \item \textbf{Note:} Values are not precisely the ones from Caen et al.~\cite{caen_reconstitution_2019} and do not consider std errors. They are derived values to be simple for this proof of concept.
\end{tablenotes}
\end{table}

\section{The Evolutionary Algorithm}
\label{app:grid_search_algorithm_parameters}

\begin{figure}
    \centering
    \captionsetup{width=0.6\linewidth}
    \includegraphics[width=0.6\linewidth]{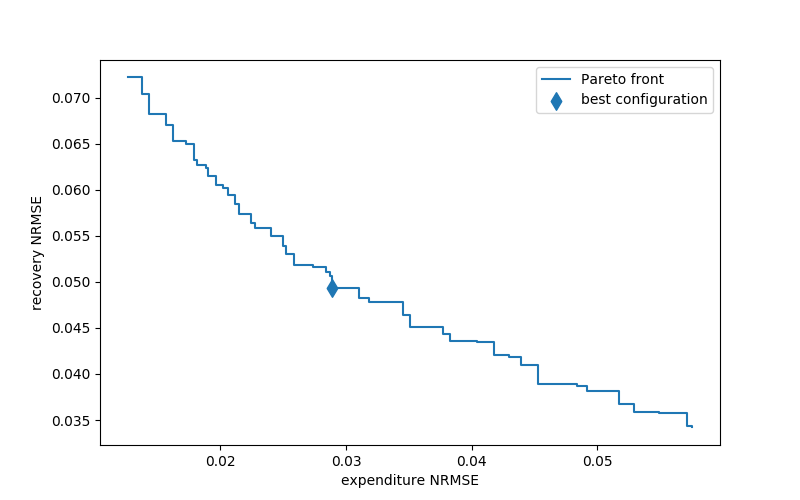}
    \caption{An exemplary Pareto front of an island. The best configuration on the Pareto front (blue diamond) is the one with minimal distance to point ($0,0$) and marks the best possible trade-off between energy expenditure and recovery error.}
    \label{fig:pareto_front}
\end{figure}

A configuration $c$ consists of eight real-valued parameters. A successful strategy to approximate an optimal configuration in such a search space is evolutionary computation~\cite{eiben_introduction_2015, back_overview_1993, biscani_parallel_2020}. The intention of this work is to provide a proof of concept that the proposed hydraulic model can be fitted to an athlete. Since a proof of concept is enough, the algorithm and parameter choices are not throughout fine tuned and we want to emphasize that much more room for parameter optimization and exploration of problem-specific algorithms is left for future work

The defined objective functions evaluate expenditure and recovery as two distinct objectives, we chose the established Multi-Objective Evolutionary Algorithm with Decomposition (MOEA/D) approach as implemented in Pygmo~\cite{biscani_parallel_2020, qingfu_zhang_moead_2007}. 

Two objectives are to be minimized: energy expenditure and recovery. 
Both objectives return a normalized root mean squared error of twelve measurements and are directly comparable. That allows to define the best trade-off between both dynamics to be the configuration on the Pareto front that has the smallest Euclidean distance to point $(0,0)$ (See example in ~\Cref{fig:pareto_front}).

In initial experiments we obtained solutions in two categories: One heavily focused on the optimization of expenditure dynamics and the other one on the recovery dynamics. To improve generalization and to consistently find configurations that optimize both objectives, we couple MOEA/D with the asynchronous islands functionality of Pygmo~\cite{biscani_parallel_2020}. That means several instances (one for each island) of the evolutionary algorithm are run isolated from each other. After a set number of generations, solutions from each of the island populations travel in-between islands. Then each algorithm continues to evolve their population, which now contains a few migrant solutions from the populations of the other algorithms. This step of evolving for a set number of generations and then exchanging solutions will be referred to as a cycle. 

Except for the number of generations, population sizes, the number of islands, and the number of cycles, all parameters are at the default that Pygmo provides. Default parameters taken from Pygmo are summarized in \Cref{tab:default_moead}.

\begin{table}
\centering
\caption{Pygmo~\cite{biscani_parallel_2020} default parameters for MOEA/D}
\label{tab:default_moead}
\begin{tabular}{ l|l }
 parameter & value  \\
 \hline
    \texttt{weight\_generation} &  'grid' \\
     \texttt{decomposition} & 'tchebycheff' \\
     \texttt{neighbors} & 20 \\
     \texttt{CR} & 1 \\
     \texttt{F} & 0.5 \\
     \texttt{eta\_m} & 20\\
     \texttt{realb} & 0.9\\
     \texttt{limit} & 2\\
     \texttt{preserve\_diversity} & true\\
     \texttt{seed} & random
\end{tabular}
\end{table}

Also for the asynchronous island approach we chose the default parameters of Pygmo in terms of migration types ('p2p') between islands and typologies ('fully connected'), but we do investigate various combinations of cycles (10,40,80), generations (10,20,30), population sizes (32,64), and numbers of islands (7,14,21). If the overall best fitness value of all island populations is not improved for more than 10 cycles, computations are stopped and the best solution of the last cycle is the returned result. The algorithm is run for ten times with each of the resulting parameter combinations. Results of all individual runs are summarized with the best (min), average (mean), and worst (max) distance result in \Cref{tab:grid_search}. The configuration of the hydraulic model that resulted in the minimal distance (best configuration) is also denoted to give an idea of how sensitive the objectives are to changes in the 8 variables of a configuration.

\begin{table*}
\begin{adjustwidth}{-.5in}{-.5in}  
\centering
\caption{Grid search. Each parameter combination was run for 10 times.}
\label{tab:grid_search}
{\small
\begin{tabular}{lllllllllllllll}
\hline
\multicolumn{7}{c|}{} & 
\multicolumn{8}{c}{resulting best configuration parameters} \\
\hline
 \multicolumn{1}{l}{gens} & \multicolumn{1}{l}{cycles} & \multicolumn{1}{l}{pop} & \multicolumn{1}{l}{islands} & \multicolumn{1}{l}{min} & \multicolumn{1}{l}{average} & \multicolumn{1}{l|}{max} & \multicolumn{1}{l}{$AnF$} & \multicolumn{1}{l}{$AnS$} & \multicolumn{1}{l}{$m^O$} & \multicolumn{1}{l}{$m^{AnS}$} & \multicolumn{1}{l}{$m^{AnF}$} & \multicolumn{1}{l}{$\phi$} & \multicolumn{1}{l}{$\theta$} & \multicolumn{1}{l}{$\gamma$} \\
 \hline
10 & 10 & 32 & 7 & 0.0750 & 0.0896 & 0.1129 & 14887.19 & 78441.50 & 247.88 & 91.73 & 10.02 & 0.64 & 0.21 & 0.32  \\ 
10 & 10 & 32 & 14 & 0.0732 & 0.0807 & 0.1020 & 16731.79 & 48023.39 & 246.03 & 113.48 & 9.60 & 0.70 & 0.01 & 0.29  \\ 
10 & 10 & 32 & 21 & 0.0698 & 0.0766 & 0.0826 & 19526.85 & 78133.12 & 247.88 & 96.16 & 8.06 & 0.79 & 0.05 & 0.19  \\ 
10 & 10 & 64 & 7 & 0.0707 & 0.0795 & 0.1028 & 18628.31 & 56881.67 & 247.51 & 108.59 & 8.15 & 0.76 & 0.02 & 0.22  \\ 
10 & 10 & 64 & 14 & 0.0710 & 0.0730 & 0.0806 & 18325.37 & 53462.06 & 248.24 & 96.52 & 8.12 & 0.72 & 0.06 & 0.22  \\ 
10 & 10 & 64 & 21 & 0.0704 & 0.0720 & 0.0756 & 18496.46 & 86583.43 & 248.30 & 91.07 & 8.91 & 0.78 & 0.09 & 0.20  \\ 
10 & 40 & 32 & 7 & 0.0703 & 0.0745 & 0.0822 & 19074.54 & 255788.43 & 248.38 & 84.80 & 8.86 & 0.83 & 0.13 & 0.18  \\ 
10 & 40 & 32 & 14 & 0.0705 & 0.0734 & 0.0894 & 18985.98 & 138678.84 & 248.24 & 85.07 & 8.09 & 0.80 & 0.12 & 0.19  \\ 
10 & 40 & 32 & 21 & 0.0697 & 0.0707 & 0.0721 & 18664.83 & 94435.01 & 247.49 & 92.78 & 8.97 & 0.79 & 0.08 & 0.20  \\ 
10 & 40 & 64 & 7 & 0.0695 & 0.0708 & 0.0749 & 19035.04 & 78738.60 & 248.04 & 94.84 & 8.08 & 0.78 & 0.07 & 0.20  \\ 
10 & 40 & 64 & 14 & 0.0697 & 0.0704 & 0.0716 & 18082.63 & 47905.06 & 247.52 & 108.79 & 9.07 & 0.72 & 0.01 & 0.22  \\ 
10 & 40 & 64 & 21 & 0.0694 & 0.0699 & 0.0703 & 19324.12 & 222108.58 & 247.90 & 81.11 & 9.01 & 0.81 & 0.14 & 0.18  \\ 
10 & 80 & 32 & 7 & 0.0706 & 0.0737 & 0.0846 & 19061.94 & 98334.88 & 247.75 & 94.97 & 8.72 & 0.82 & 0.05 & 0.19  \\ 
10 & 80 & 32 & 14 & 0.0698 & 0.0723 & 0.0769 & 17861.85 & 86634.11 & 247.95 & 90.63 & 9.53 & 0.75 & 0.11 & 0.22  \\ 
10 & 80 & 32 & 21 & 0.0699 & 0.0713 & 0.0769 & 18917.47 & 144408.28 & 247.90 & 86.06 & 8.52 & 0.80 & 0.12 & 0.20  \\ 
10 & 80 & 64 & 7 & 0.0698 & 0.0707 & 0.0751 & 19031.40 & 119081.83 & 248.05 & 85.09 & 8.81 & 0.79 & 0.11 & 0.19  \\ 
10 & 80 & 64 & 14 & 0.0696 & 0.0706 & 0.0750 & 17958.82 & 67994.85 & 247.73 & 95.95 & 9.01 & 0.74 & 0.08 & 0.23  \\ 
10 & 80 & 64 & 21 & 0.0697 & 0.0699 & 0.0704 & 17725.86 & 50325.31 & 247.26 & 107.09 & 9.27 & 0.72 & 0.02 & 0.23  \\ 
20 & 10 & 32 & 7 & 0.0702 & 0.0775 & 0.1011 & 18922.50 & 69263.96 & 248.17 & 94.35 & 8.45 & 0.76 & 0.06 & 0.19  \\ 
20 & 10 & 32 & 14 & 0.0709 & 0.0742 & 0.0805 & 17021.40 & 46888.56 & 247.03 & 106.46 & 9.81 & 0.68 & 0.04 & 0.26  \\ 
20 & 10 & 32 & 21 & 0.0707 & 0.0731 & 0.0754 & 20195.32 & 101888.06 & 248.22 & 91.04 & 7.81 & 0.82 & 0.06 & 0.17  \\ 
20 & 10 & 64 & 7 & 0.0700 & 0.0728 & 0.0768 & 18925.49 & 178845.77 & 248.02 & 85.65 & 9.63 & 0.80 & 0.13 & 0.19  \\ 
20 & 10 & 64 & 14 & 0.0699 & 0.0703 & 0.0707 & 17642.03 & 79591.96 & 247.83 & 95.06 & 9.30 & 0.75 & 0.10 & 0.23  \\ 
20 & 10 & 64 & 21 & 0.0695 & 0.0706 & 0.0725 & 18063.94 & 89567.17 & 247.71 & 91.83 & 9.48 & 0.75 & 0.11 & 0.22  \\ 
20 & 40 & 32 & 7 & 0.0704 & 0.0745 & 0.0966 & 18240.97 & 62161.99 & 248.48 & 95.40 & 8.83 & 0.74 & 0.06 & 0.20  \\ 
20 & 40 & 32 & 14 & 0.0701 & 0.0717 & 0.0742 & 20993.06 & 98680.10 & 248.57 & 86.20 & 7.18 & 0.82 & 0.07 & 0.15  \\ 
20 & 40 & 32 & 21 & 0.0704 & 0.0713 & 0.0721 & 19882.77 & 132015.35 & 248.53 & 85.79 & 8.22 & 0.81 & 0.10 & 0.17  \\ 
20 & 40 & 64 & 7 & 0.0695 & 0.0702 & 0.0734 & 17553.77 & 75082.95 & 247.60 & 91.18 & 9.70 & 0.73 & 0.11 & 0.23  \\ 
20 & 40 & 64 & 14 & 0.0694 & 0.0698 & 0.0704 & 18287.25 & 50128.61 & 247.41 & 103.44 & 9.04 & 0.72 & 0.02 & 0.22  \\ 
20 & 40 & 64 & 21 & 0.0691 & 0.0696 & 0.0698 & 18217.42 & 175251.33 & 248.05 & 85.18 & 9.26 & 0.78 & 0.15 & 0.21  \\ 
20 & 80 & 32 & 7 & 0.0697 & 0.0713 & 0.0737 & 19431.50 & 160339.17 & 248.28 & 83.54 & 8.76 & 0.80 & 0.13 & 0.19  \\ 
20 & 80 & 32 & 14 & 0.0700 & 0.0706 & 0.0740 & 18631.83 & 61995.02 & 247.59 & 99.87 & 8.95 & 0.76 & 0.04 & 0.21  \\ 
20 & 80 & 32 & 21 & 0.0700 & 0.0705 & 0.0718 & 17613.36 & 56808.61 & 247.58 & 96.82 & 9.45 & 0.71 & 0.07 & 0.23  \\ 
20 & 80 & 64 & 7 & 0.0695 & 0.0699 & 0.0704 & 19706.74 & 83704.90 & 248.09 & 93.96 & 8.12 & 0.80 & 0.06 & 0.18  \\ 
20 & 80 & 64 & 14 & 0.0694 & 0.0698 & 0.0701 & 19214.76 & 65330.19 & 247.50 & 100.84 & 8.43 & 0.78 & 0.03 & 0.20  \\ 
20 & 80 & 64 & 21 & 0.0695 & 0.0697 & 0.0701 & 18708.34 & 98318.64 & 247.66 & 90.08 & 9.14 & 0.78 & 0.10 & 0.20  \\ 
30 & 10 & 32 & 7 & 0.0720 & 0.0782 & 0.0903 & 17042.00 & 185592.26 & 247.77 & 84.21 & 9.24 & 0.74 & 0.19 & 0.25  \\ 
30 & 10 & 32 & 14 & 0.0701 & 0.0723 & 0.0753 & 18033.01 & 161795.47 & 247.53 & 85.22 & 9.75 & 0.78 & 0.14 & 0.21  \\ 
30 & 10 & 32 & 21 & 0.0702 & 0.0715 & 0.0729 & 19264.53 & 77426.94 & 248.31 & 92.80 & 8.44 & 0.78 & 0.07 & 0.19  \\ 
30 & 10 & 64 & 7 & 0.0694 & 0.0707 & 0.0733 & 19155.61 & 58966.01 & 248.05 & 101.65 & 7.88 & 0.77 & 0.03 & 0.20  \\ 
30 & 10 & 64 & 14 & 0.0693 & 0.0700 & 0.0706 & 19544.79 & 70406.66 & 248.04 & 94.11 & 8.36 & 0.78 & 0.05 & 0.18  \\ 
30 & 10 & 64 & 21 & 0.0693 & 0.0700 & 0.0711 & 18348.18 & 120883.19 & 247.66 & 85.21 & 9.11 & 0.77 & 0.13 & 0.21  \\ 
30 & 40 & 32 & 7 & 0.0701 & 0.0720 & 0.0768 & 19898.17 & 136898.73 & 248.51 & 83.87 & 8.23 & 0.81 & 0.10 & 0.17  \\ 
30 & 40 & 32 & 14 & 0.0700 & 0.0707 & 0.0715 & 18143.07 & 47370.24 & 247.84 & 105.76 & 9.07 & 0.71 & 0.02 & 0.21  \\ 
30 & 40 & 32 & 21 & 0.0698 & 0.0702 & 0.0706 & 19269.44 & 57468.40 & 247.76 & 106.11 & 8.14 & 0.77 & 0.01 & 0.20  \\ 
30 & 40 & 64 & 7 & 0.0691 & 0.0699 & 0.0716 & 18299.38 & 159976.90 & 247.83 & 84.20 & 9.24 & 0.78 & 0.15 & 0.21  \\ 
30 & 40 & 64 & 14 & 0.0692 & 0.0695 & 0.0701 & 18581.79 & 109451.44 & 247.75 & 88.95 & 9.24 & 0.78 & 0.11 & 0.20  \\ 
30 & 40 & 64 & 21 & 0.0692 & 0.0696 & 0.0702 & 19164.84 & 155545.77 & 248.26 & 83.53 & 8.15 & 0.79 & 0.14 & 0.20  \\ 
30 & 80 & 32 & 7 & 0.0693 & 0.0715 & 0.0776 & 18397.78 & 48646.16 & 247.98 & 103.95 & 8.62 & 0.72 & 0.02 & 0.22  \\ 
30 & 80 & 32 & 14 & 0.0701 & 0.0702 & 0.0704 & 17629.46 & 51157.57 & 247.16 & 106.93 & 9.23 & 0.72 & 0.03 & 0.24  \\ 
30 & 80 & 32 & 21 & 0.0694 & 0.0702 & 0.0716 & 19211.40 & 75566.60 & 247.95 & 96.18 & 8.26 & 0.79 & 0.05 & 0.19  \\ 
30 & 80 & 64 & 7 & 0.0695 & 0.0698 & 0.0702 & 18858.54 & 56032.86 & 247.37 & 103.52 & 8.99 & 0.75 & 0.02 & 0.20  \\ 
30 & 80 & 64 & 14 & 0.0693 & 0.0696 & 0.0699 & 18295.58 & 115731.00 & 247.62 & 86.74 & 9.15 & 0.77 & 0.12 & 0.21  \\ 
30 & 80 & 64 & 21 & 0.0692 & 0.0696 & 0.0699 & 18645.05 & 260070.20 & 248.17 & 81.43 & 8.96 & 0.79 & 0.16 & 0.20  \\ 
\end{tabular}
}
\end{adjustwidth}
\end{table*}

\clearpage

\bibliographystyle{unsrt}  

\begin{thebibliography}{10}

\bibitem{clarke_rationale_2013}
David~C. Clarke and Philip~F. Skiba.
\newblock Rationale and resources for teaching the mathematical modeling of
  athletic training and performance.
\newblock {\em Advances in Physiology Education}, 37(2):134--152, June 2013.

\bibitem{jones_critical_2017}
Andrew~M. Jones and Anni Vanhatalo.
\newblock The 'critical power' concept: applications to sports performance with
  a focus on intermittent high-intensity exercise.
\newblock {\em Sports Medicine}, 47(S1):65--78, March 2017.

\bibitem{sreedhara_survey_2019}
Vijay Sarthy~M. Sreedhara, Gregory~M. Mocko, and Randolph~E. Hutchison.
\newblock A survey of mathematical models of human performance using power and
  energy.
\newblock {\em Sports Medicine - Open}, 5(1):54, December 2019.

\bibitem{caen_reconstitution_2019}
Kevin Caen, Jan~G. Bourgois, Gil Bourgois, Thibaux Van Der~Stede, Kobe
  Vermeire, and Jan Boone.
\newblock The reconstitution of {W}' depends on both work and recovery
  characteristics.
\newblock {\em Medicine \& Science in Sports \& Exercise}, 51(8):1745--1751,
  August 2019.

\bibitem{morton_three_1986}
R.~Hugh Morton.
\newblock A three component model of human bioenergetics.
\newblock {\em Journal of Mathematical Biology}, 24(4):451--466, July 1986.

\bibitem{morton_critical_2006}
R.~Hugh Morton.
\newblock The critical power and related whole-body bioenergetic models.
\newblock {\em European Journal of Applied Physiology}, 96(4):339--354, March
  2006.

\bibitem{biscani_parallel_2020}
Francesco Biscani and Dario Izzo.
\newblock A parallel global multiobjective framework for optimization: pagmo.
\newblock {\em Journal of Open Source Software}, 5(53):2338, September 2020.

\bibitem{morton_modelling_1990}
R.~Hugh Morton.
\newblock Modelling human power and endurance.
\newblock {\em Journal of Mathematical Biology}, 28(1):49--64, January 1990.

\bibitem{morton_model_1985}
R.~Hugh Morton.
\newblock On a model of human bioenergetics.
\newblock {\em European Journal of Applied Physiology and Occupational
  Physiology}, 54(3):285--290, September 1985.

\bibitem{morton_model_1986}
R.~Hugh Morton.
\newblock On a model of human bioenergetics {II}: maximal power and endurance.
\newblock {\em European Journal of Applied Physiology and Occupational
  Physiology}, 55(4):413--418, August 1986.

\bibitem{sundstrom_comparing_2014}
David Sundström, Peter Carlsson, and Mats Tinnsten.
\newblock Comparing bioenergetic models for the optimisation of pacing strategy
  in road cycling.
\newblock {\em Sports Engineering}, 17(4):207--215, December 2014.

\bibitem{sundstrom_optimization_2013}
David Sundström, Peter Carlsson, and Mats Tinnsten.
\newblock On optimization of pacing strategy in road cycling.
\newblock {\em Procedia Engineering}, 60:118--123, 2013.

\bibitem{hill_critical_1993}
David~W. Hill.
\newblock The critical power concept: a review.
\newblock {\em Sports Medicine}, 16(4):237--254, October 1993.

\bibitem{eiben_introduction_2015}
A.E. Eiben and J.E. Smith.
\newblock {\em Introduction to evolutionary computing}.
\newblock Natural {Computing} {Series}. Springer Berlin Heidelberg, Berlin,
  Heidelberg, 2015.

\bibitem{back_overview_1993}
Thomas Bäck and Hans-Paul Schwefel.
\newblock An overview of evolutionary algorithms for parameter optimization.
\newblock {\em Evolutionary Computation}, 1(1):1--23, March 1993.

\bibitem{qingfu_zhang_moead_2007}
{Qingfu Zhang} and {Hui Li}.
\newblock {MOEA}/{D}: {A} {Multiobjective} {Evolutionary} {Algorithm} {Based}
  on {Decomposition}.
\newblock {\em IEEE Transactions on Evolutionary Computation}, 11(6):712--731,
  December 2007.

\end{thebibliography}

\end{document}